\documentclass{article}

\usepackage{microtype}
\usepackage{graphicx}
\usepackage{subfigure}
\usepackage{booktabs} 

\usepackage{hyperref}

\usepackage[accepted]{icml2023}

\usepackage{amsmath}
\usepackage{amssymb}
\usepackage{mathtools}
\usepackage{amsthm}
\usepackage{bbm}
\usepackage{multirow}
\usepackage{makecell}
\usepackage[capitalize,noabbrev]{cleveref}

\theoremstyle{plain}

\theoremstyle{definition}

\theoremstyle{remark}

\usepackage[textsize=tiny]{todonotes}

\begin{document}

\twocolumn[
\icmltitle{Boosting Multi-modal Model Performance with Adaptive Gradient Modulation}



\icmlsetsymbol{equal}{*}

\begin{icmlauthorlist}
\icmlauthor{Hong Li $^*$}{shanghaiTech,shic}
\icmlauthor{Xingyu Li $^*$}{bibii}
\icmlauthor{Pengbo Hu}{ustc-sist}
\icmlauthor{Yinuo Lei}{shanghaiTech,shic}
\icmlauthor{Chunxiao Li}{ustc-management}
\icmlauthor{Yi Zhou}{ustc-sist,ustc-nel,ustc-klbipc}
\end{icmlauthorlist}

\icmlaffiliation{shanghaiTech}{School of Information Science and Technology, ShanghaiTech University}
\icmlaffiliation{bibii}{Shanghai Center for Brain Science and Brain-Inspired Technology, Shanghai, China}
\icmlaffiliation{ustc-sist}{School of Information Science and Technology, University of Science and Technology of China, Hefei, China}
\icmlaffiliation{ustc-nel}{National Engineering Laboratory for Brain-inspired Intelligence Technology and Application, University of Science and Technology of China, Hefei, China}
\icmlaffiliation{ustc-klbipc}{Key Laboratory of Brain-inspired Intelligent Perception and Cognition (University of Science and Technology of China), Ministry of Education}
\icmlaffiliation{ustc-management}{School of Management, University of Science and Technology of China, Hefei, China}
\icmlaffiliation{shic}{Shanghai Innovation Center for Processor Technologies}

\icmlcorrespondingauthor{Yi Zhou}{yi\_zhou@ustc.edu.cn}


\vskip 0.3in
]



\printAffiliationsAndNotice{\icmlEqualContribution} 

\begin{abstract}
While the field of multi-modal learning keeps growing fast, the deficiency of the standard joint training paradigm has become clear through recent studies. They attribute the sub-optimal performance of the jointly trained model to the modality competition phenomenon. Existing works attempt to improve the jointly trained model by modulating the training process. Despite their effectiveness, those methods can only apply to late fusion models. More importantly, the mechanism of the modality competition remains unexplored. In this paper, we first propose an adaptive gradient modulation method that can boost the performance of multi-modal models with various fusion strategies. Extensive experiments show that our method surpasses all existing modulation methods. Furthermore, to have a quantitative understanding of the modality competition and the mechanism behind the effectiveness of our modulation method, we introduce a novel metric to measure the competition strength. This metric is built on the mono-modal concept, a function that is designed to represent the competition-less state of a modality. Through systematic investigation, our results confirm the intuition that the modulation encourages the model to rely on the more informative modality. In addition, we find that the jointly trained model typically has a preferred modality on which the competition is weaker than other modalities. However, this preferred modality need not dominate others. Our code will be available at \url{https://github.com/lihong2303/AGM_ICCV2023}.
\end{abstract}

\begin{figure*}
\begin{center}
\includegraphics[width=0.9\linewidth]{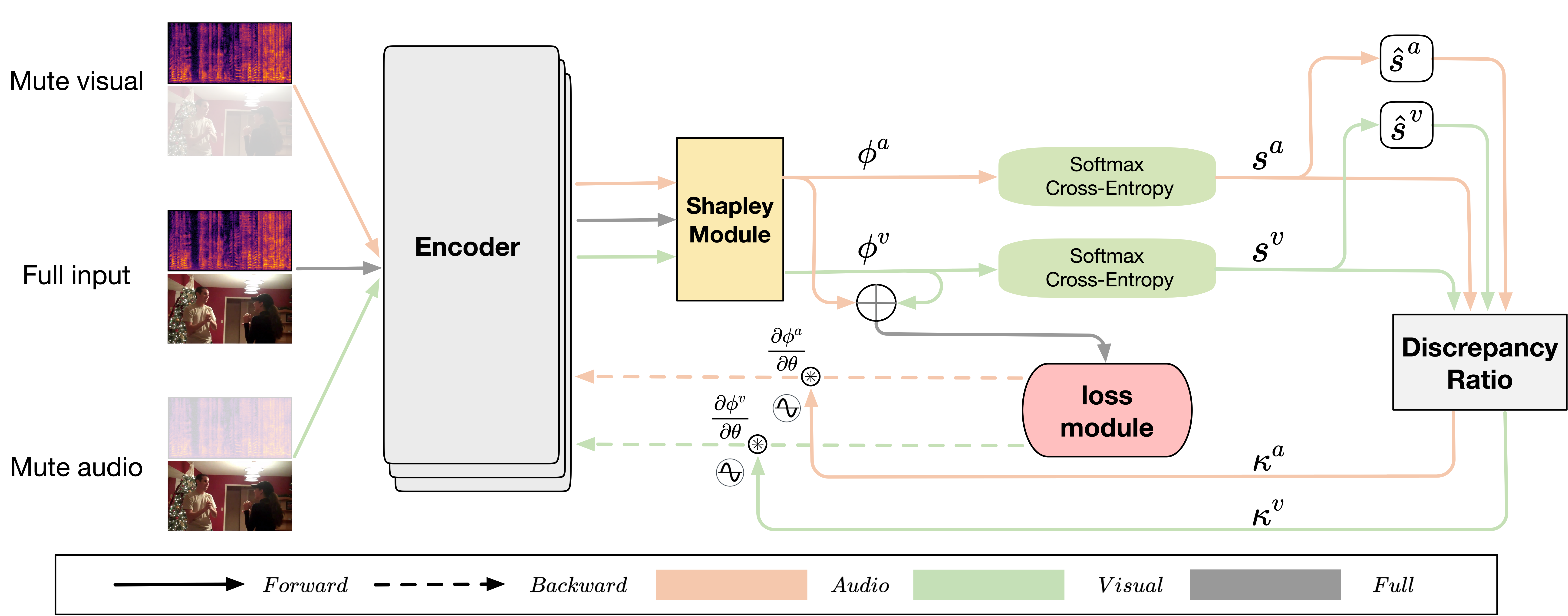}
\end{center}
   \caption{Schematic diagram of the adaptive gradient modulation (AGM) method. 
   Firstly, based on the full input and corresponding muted inputs, the Shapley module produces mono-modal outputs $\phi^{m}$, which disentangle the responses of the multi-modal model to individual modalities.
   Next, $\phi^{m}$ are used to compute the mono-modal cross-entropy $s^{m}$ that reflects the amount of information in modality $m$. 
   At last, $s^{m}$ and their running average $\hat{s}^{m}$ are fed to the Discrepancy Ratio module to compute the modulation coefficients $\kappa^{m}$ for each modality, which in turn modulate the strength of corresponding gradient signals during back-propagation.}
\label{fig:teaser-figure}
\end{figure*}

\section{Introduction}

Recent years have seen tremendous progress in deep multi-modal learning. Despite these advances, integrating information from multiple modalities remains challenging. Many efforts have been made to design sophisticated fusion methods for better performance. However, adding additional modalities only slightly improves accuracy in some multi-modal tasks.  For example, trained on the CMU-MOSEI \cite{delbrouck2020transformer} dataset, the accuracy of the text-based single-modal model is only about $1\%$ point lower than that of the multi-modal model based on both text and audio modalities. Similar phenomena have also been observed across a wide variety of multi-modal datasets~\cite{vielzeuf2018centralnet,cao2014crema}.

Such inefficiency in exploiting and integrating information from multiple modalities presents a great challenge to the multi-modal learning field.
It is commonly believed that this inefficiency is a consequence of the existence of the dominant modality, which prevents the model from fully exploiting the other relatively weak modalities~\cite{ma2022multimodal,huSHAPEUnifiedApproach2022}.
Recent studies~\cite{allen2020towards,huangModalityCompetitionWhat2022,han2022trusted} theoretically investigate the training process of late fusion models and explain the production of the dominant modality with the concept of modality competition. In addition to the theoretical studies, there is a group of empirical works that attempts to develop methods to modulate the training of a multi-modal model and balance the learning of different modalities and, thus, achieve better performance.
To our best knowledge, existing modulation methods are confined to late fusion models which greatly limits their application. More importantly, little effort has been paid to the study of the mechanism behind the effectiveness of those modulation methods.

It is natural to ask \textit{Can we design a modulation method that applies to more complex fusion strategies?} and \textit{Is it possible to understand the working mechanism of modulation in terms of modality competition?} To this end, we propose an adaptive gradient modulation method, which utilizes a Shapley value-based attribution technique, that can in principle apply to any fusion strategy. Our approach achieves better performance compared with the current modulation methods. 
Moreover, we introduce the mono-modal concept to represent the competition-less state of a modality and build a metric on top of it to directly measure the competition strength of a modality in a multi-modal model. This novel metric lay the base for us to quantitatively study the behavior of modality competition and the working mechanism of our adaptive gradient modulation method.

Our main contributions are three-fold:
\begin{itemize}
\item[1.] We propose an adaptive gradient modulation method that can boost the performance of multi-modal models with various fusion strategies and justify its effectiveness through extensive experiments.
\item[2.] We introduce the mono-modal concept to capture the competition-less state of a modality and build a novel metric to measure the modality competition strength.
\item[3.] We systematically analyze the behavior of modality competition and study the mechanism of how our modulation method works.
\end{itemize}

\section{Related work}

\subsection{Multi-modal learning}
Multi-modal learning is a fast-growing research area. 
It addresses the needs of effectively processing multi-sensory data in real-world tasks and has applications in various fields, such as 
multi-modal sentiment classification~\cite{zadeh2018multimodal,cao2014crema}, audio-visual localization~\cite{tian2018audio}
and visual question answering~\cite{antol2015vqa,ilievski2017multimodal,wu2021star}. 
According to the fusion strategy, one distinguishes three types~\cite{baltrusaitisMultimodalMachineLearning2017}, i.e., the late fusion, the early fusion, and the hybrid fusion, depending on when the fusion happens at the output stage, at the input stage, and in a complex manner, respectively. 
From another perspective, 
existing models can be divided into two categories, either jointly training different modalities in an end-to-end fashion or exploiting pre-trained models and building a multi-stage pipeline.

In this paper, we focus on the multi-modal joint training models for the multi-modal classification task, and we will compare models with different fusion strategies.

\subsection{Modality-specific modulation} 

Recent studies~\cite{wangWhatMakesTraining2020, huangModalityCompetitionWhat2022}
reveal the deficiency of the multi-modal joint training paradigm 
that information on the input modalities is often under-exploited. 
To address this deficiency, existing works commonly propose to intervene in the training process.
~\citet{geng2021uncertainty}
propose to obtain noise-free multi-view representations with the help of uncertainty in Dynamic Uncertainty-Aware Networks.
~\citet{wang2020makes}
devise the Gradient-blending technique which addresses the overfitting in a multi-modal model by optimally blending modalities.
~\citet{wu2022characterizing}
propose to balance the speed of learning from different modalities based on their conditional utilization rates.
~\citet{fujimoriModalitySpecificLearningRate2020}
emphasize the heterogeneity
of different network branches in joint training and propose to avoid overfitting through modality-specific early stopping.
~\citet{yaoModalityspecificLearningRates2022} advocates using modality-specific learning rates for different branches in a multi-modal model to fully explore the capacity of the corresponding network architecture.
More recently, ~\citet{pengBalancedMultimodalLearning2022a} proposes to adjust the gradients of individual modalities based on their output magnitudes.
The assumption is that in an ideal multi-modal model, the outputs of individual modalities should be balanced, i.e., having similar magnitudes.
Consequently, the gradient of the modality with larger outputs will be modulated on-the-fly towards a lower magnitude during each training iteration.

Despite the effectiveness of the above-mentioned methods, 
they are all confined to late fusion models, limiting their practical use.
More importantly, the mechanism of why those methods work to improve the multi-modal model
remains unexplored.

\subsection{Mono-modal behavior}

One way to investigate the mechanism underlying a multi-modal model is to quantify how much modalities affect each other in the model. 
In a recent theoretical analysis, ~\citet{huangModalityCompetitionWhat2022} term this interaction among modalities as the modality competition.

Due to the complexity and non-linearity of neural network models, 
it is infeasible to isolate a part of the computations that account for the competition.
Existing works instead attempt to measure the mono-modal behavior inside a multi-modal model, 
which can partly reflect the interactions among modalities.
~\citet{hesselDoesMyMultimodal2020a} design the empirical multimodally-additive function projection (EMAP) that implicitly reflects the mono-modal behavior by averaging out all other modalities.
\citet{yaoModalityspecificLearningRates2022} employ the layer conductance~\cite{shrikumarComputationallyEfficientMeasures2018b} to evaluate the importance of individual modalities in late fusion models. 
~\citet{gatPerceptualScoreWhat2021a} propose the perceptual scores to measure the mono-modal importance directly. The key idea of their method is the input permutation, 
which removes the influence of modalities other than the targeting one.
What is most related to the goal of measuring the modality competition is the recently proposed SHAPE scores~\cite{huSHAPEUnifiedApproach2022}. 
The authors devise a way to compute the cross-modal cooperation strength based on the Shapley values.

It is worth noting that all the above-mentioned methods are self-oriented in the sense that 
they only utilized the multi-modal model, where competition already presents.
The lack of information about how each modality behaves without competition prevents those models from faithfully reflecting the modality competition strength.

\section{Method}

\subsection{Adaptive gradient modulation}

Drawing inspiration from the Shapley value-based attribution method~\cite{huSHAPEUnifiedApproach2022} and the On-the-fly gradient modulation generalization enhancement (OGM-GE) algorithm~\cite{pengBalancedMultimodalLearning2022a}, we propose an adaptive gradient modulation (AGM) method that modulates the level of participation of individual modalities. 
\Cref{fig:teaser-figure} presents the illustration of the proposed AGM.
Our approach is in line with the OGM-GE algorithm in the sense that
both attempt to balance the mono-modal responses in a multi-modal model. 

Nonetheless, our approach differs from the OGM-GE in the following three important aspects:
1) We adopt a Shapley value-related method to compute the mono-modal responses. In this way, our approach applies to complex fusion strategies rather than being limited to the late fusion case.
2) We extend the method to calculate the discrepancy ratios so that our approach can deal with situations with more than two modalities.
3) In our approach, the discrepancy ratios are modulated towards their running average rather than $1$, reflecting the distinctions among different modalities.

\subsubsection{Isolating the mono-modal responses}

The core component of our approach is the algorithm to isolate the mono-modal responses, 
which enables us to further compute the mono-modal cross entropy and the mono-modal accuracy.

Let $\phi(x), x=(x^{m_1},\dots,x^{m_k})$ be a multi-modal model on the data with $k$ modalities and $\mathcal{M}:=\{m_i\}_{i\in[k]}$ be the set of all modalities. 
Same as in~\cite{huSHAPEUnifiedApproach2022} we use zero-padding $0^{m}$ to represent the absence of features of modality $m$. 
When $S$ is a subset of $\mathcal{M}$, $\phi(S)$ denotes that if $m\in S$, the component $x^m$ is substituted with $0^m$.
Then the mono-modal response for $m$ is defined as 
\begin{equation}\label{eq:monomodalresp}
    \phi^m(x) = \sum_{S\subseteq \mathcal{M}/\{m\}; S\ne \emptyset}
    \frac{|S|!(k-|S|-1)!}{k!} V_m(S;\phi),
\end{equation}
where $V_m(S;\phi) = \phi(S\cup \{m\}) - \phi(S)$. 
Note that we exclude the empty subset from the above summation.
In this way, we ensure the relation 
\begin{equation}
\phi(x)=\sum_m \phi^m(x).    
\end{equation}

As an example, for the two-modality case~\cref{eq:monomodalresp} is simplified to
\begin{equation}
\begin{split}
    \phi^{m_1}(x) = \frac{1}{2}
    &\left[ \phi(\{m_1,m_2\}) - \phi(\{0^{m_1},m_2\}) \right. \\ 
    &\left.+ \phi(\{m_1, 0^{m_2}\}) \right].
\end{split}
\end{equation}

The mono-modal cross entropy and mono-modal accuracy are then defined subsequently,
\begin{equation}
    s^m = \mathbb{E}_{x\sim \mathcal{D}}\left[
    -\log{\left( \text{Softmax}(\phi^m(x))_y \right)}
    \right],
\end{equation}
and
\begin{equation}
    Acc_m = \mathbb{E}_{x\sim \mathcal{D}}\left[
    \mathbbm{1}_{y=y_p(x)}
    \right],
\end{equation}
where $y$ is the ground-truth class of $x$ and $y_p$ the model prediction,
$y_p(x)=\arg\max_{y'\in[K]} \phi^m_{y'}(x)$.

\subsubsection{Modulating the training process}
\label{sec: Gradient Adaptive Control}

We modulate the level of participation of individual modalities through 
adjusting the intensity of the back-propagation signal of each modality,
\begin{equation}
    \theta_{t+1} = \theta_t - 
    \eta \frac{\partial \mathcal{L}}{\partial \phi} \cdot 
    \sum_m \kappa^m_t \frac{\partial \phi^m}{\partial \theta}\bigg\rvert_{t},
\end{equation}
where $t$ refers to a specific iteration of training, $\theta$ denotes the trainable network parameters, $\eta$ is the learning rate and $\mathcal{L}$ is the loss function. 

Coefficient $\kappa^m_t$ controls the magnitude of the update signal for modality $m$ at iteration $t$. Intuitively, if a modality is too strong (weak) we want to suppress (amplify) its update signal. The strength of a modality is measured by the averaged differences relative to the other modalities
\begin{equation}
    r^m_t = \exp{\left(\frac{1}{K-1}\sum_{m'\in[K];m'\ne m}(s^m_t - s^{m'}_t)\right)}.
\end{equation}
We choose to compare different modalities based on their mono-modal cross-entropy, 
since $s^m_t$ reflects the amount of information attributed to modality $m$ within the full model outputs.
Then $\kappa^m_t$ is defined as follows
\begin{equation}
    \kappa^m_t = \exp{\left( -\alpha*\left( r^m_t - \tau^m_t \right) \right)},
\end{equation}
where $\alpha>0$ is a hyper-parameter that controls the degree of modulation and 
$\tau^m_t$ is the reference for modulation.
Consequently, when a modality is too strong ($ r^m_t > \tau^m_t$), we lower its update signal ($\kappa^m_t < 1$).

In the current implementation, we choose $\tau^m_t$ to be
\begin{equation}
    \tau^m_t = \exp{
    \left( 
    \frac{1}{K-1}\sum_{m'\in[K];m'\ne m}\left(\hat{s}^m(t) - \hat{s}^{m'}(t) \right) 
    \right)},
\end{equation}
where $\hat{s}^m(t)$ denotes the running average of mono-modal cross-entropy at iteration $t$, 
\begin{equation}
    \hat{s}^{m}(t) = \hat{s}^{m}(t-1) \cdot \frac{t-1}{t} + \frac{s^{m}_t}{t}.
\end{equation} 
The above steps are summarized in~\Cref{alg:GAC} below.

\begin{algorithm}
    \caption{Adaptive Gradient Modulation}
    \begin{algorithmic}[1]\label{alg:GAC}
        \STATE Training dataset $\mathcal{D} = \{(x^{m_1},x^{m_2},..,x^{m_k}), y_i\}$, iteration number $T$, logits output of a modality $o^{m}_{t}$, model logits output $o_{t}$, softmax output of a modality $p^{m}_{t}$, batch size $N$, mono-modal information $s^{m}_{t}$, batch information discrepancy $r^{m}_{t}$, running average information discrepancy $\tau^{m}_{t}$, modulation coefficient $\kappa^{m}_{t}$, $m \in \{m_1, m_2,...,m_k\}$.  
        \STATE $\hat{s}^{m} = 0.$
        \FOR {t=$1,2,\ldots,T$}
            \STATE $o^{m_1}_{t},o^{m_2}_{t},...,o^{m_k}_t,o_{t} = \text{net}(x^{m_1},x^{m_2},...,x^{m_k})$
            \STATE $p^{m}_{t} = \text{Softmax}(o^{m}_{t})$
            \STATE $s^{m}_{t} = \frac{1}{N} \sum^{N}_{i=1}log^{p^{m}_{t}[i][y[i]]}$
            \STATE $\overline{s}_{t} = \frac{s_{t}^{m_{1}} + s_{t}^{m_{2}} + ,..., + s_{t}^{m_k}}{k}$, $\overline{\hat{s}}_{t} = \frac{\hat{s}_{t}^{m_{1}} + \hat{s}_{t}^{m_{2}} + ,..., + \hat{s}_{t}^{m_k}}{k}$
            \STATE $r^{m}_{t} =e^{((s^{m}_{t} - \overline{s}_{t})\cdot \frac{k}{k-1})}$, $\tau^{m}_{t} =e^{((\hat{s}^{m} - \overline{\hat{s}}_{t})\cdot\frac{k}{k-1})}$
            \STATE $\kappa^{m}_{t} = e^{(-\alpha * (r^{m}_{t} - \tau^{m}_t))}$ 
            \STATE $\hat{s}^{m} = \frac{\hat{s}^{m} \cdot t}{t+1} + \frac{s^{m}_{t}}{t+1}$
            \STATE Update using $\theta_{t+1} = \theta_t - 
    \eta \frac{\partial \mathcal{L}}{\partial \phi} \cdot 
    \sum_m \kappa^m_t \frac{\partial \phi^m}{\partial \theta}\bigg\rvert_{t}$
        \ENDFOR
    \end{algorithmic}
\end{algorithm}

\subsection{Mono-modal competition strength\label{sec: monomodal Concept}}

The empirical study~\cite{wangWhatMakesTraining2020} demonstrates that 
multi-modal joint training can lead to suboptimal performance that is even worse than the mono-modal model.
Recently, Huang et al.~\cite{huangModalityCompetitionWhat2022} theoretically study this phenomenon in a simplified setting and attribute it to the modality competition mechanism that the representation learning of a modality is generally affected by the presence of other modalities.
The authors further suggest that modality competition potentially explains the effectiveness of the adaptive learning methods~\cite{wangWhatMakesTraining2020,pengBalancedMultimodalLearning2022a},
which are designed to improve the performance of joint training.

However, the above-mentioned studies are all confined to late fusion cases. 
It remains unexplored whether the modality competition mechanism can generalize to other fusion strategies and how it alters the representation learning in realistic multi-modal models.
This leads to an urgent need for methods that directly measure competition strength.

To quantify modality competition, one must specify the competition-less state for each modality.
Previous attribution methods~\cite{hesselDoesMyMultimodal2020a,yaoModalityspecificLearningRates2022,gatPerceptualScoreWhat2021a,huSHAPEUnifiedApproach2022} only utilize the responses of the underlying multi-modal model where the competition already took place and, hence, is in principle incapable of reflecting modality competition.
To address this challenge, we introduce the mono-modal concept, which defines how the corresponding modality in a given multi-modal model will behave in the absence of modality competition.
Then the competition strength is estimated based on the deviation of the multi-modal model outputs with respect to this mono-modal concept.

\subsubsection{Mono-modal concept}

Let $x=(x^{m_1}, x^{m_2})$ denote a multi-modal input feature, 
where $x^{m_1}$ and $x^{m_2}$ refer to the mono-modal components. 
We focus on two modalities case below and the extension to more modalities is straightforward.

The processing of $x^{m_1}$ by a multi-modal model is determined by
the complementary component $x^{m_2}$,
the network architecture $\phi$\footnote{
we abuse the symbol $\phi$ a little so that it may refer to both the network architecture and the corresponding network function.
}
, the training settings $\mathcal{T}$\footnote{
$\mathcal{T}$ includes the initialization, the loss function, hyper-parameters, and specific techniques, e.g., the learning rate scheduler, used in training.
}
and the dataset $\mathcal{D}$.
We call this quadruple $\mathcal{E}_{m_1}:=(x^{m_2}, \phi, \mathcal{T}, \mathcal{D})$ as the environment of mono-modal input $x^{m_1}$.
Roughly speaking, in the competition-less state we want to remove the effects of $x^{m_2}$ while retaining the ``normal'' processing of $x^{m_1}$.
This can be formally denoted as $\mathcal{E}_{m_1}/m_2$.

With the above notations, we abstract the competition-less state for $m_1$ as a function $\mathcal{C}^{m_1}(x^{m_1}; \mathcal{E}_{m_1}/m_2)$ 
that maps the inputs to vectors in $\mathbb{R}^K$, where $K$ is the number of classes. 
Intuitively, $\mathcal{C}^{m_1}$ captures the responses, 
of a given multi-modal model,
to the mono-modal inputs without modality competition.
Following the terminology in~\cite{mcgrathAcquisitionChessKnowledge2022}, $\mathcal{C}^{m_1}$ is referred as the \emph{mono-modal concept} of modality $m_1$.
In the following, we elaborate the construction of $\mathcal{C}^m, m\in\{m_1, m_2\}$ under different situations.

\paragraph{Late fusion case.} 
In late fusion the multi-modal model can be written as $\phi(x) = \phi^{m_1}(x^{m_1}) + \phi^{m_2}(x^{m_2})$. It is natural to set $\mathcal{E}_{m_1}/m_2 = (\mathbf{0}^{m_2}, \phi^{m_1}, \mathcal{T}_{m_1}, \mathcal{D}_{m_1})$.
$\mathbf{0}^{m_2}$ denotes the null input of modality $m_2$, which is realized, in the current case, by simply discarding the branch $\phi^{m_2}$. 
$\mathcal{T}_{m_1}$ refers to the same training set for the $m_1$ branch as it was during the training of the multi-modal model $\phi$.
At last, $\mathcal{D}_{m_1}$ denotes the set of mono-model feature components $\{x^{m_1}_i\}_{i\in[N]}$,
where $N$ is the number of data samples and $[N]:=\{1,\dots,N\}$.
In practice, we need to \emph{train} $\phi^{m_1}$ on $\mathcal{D}_{m_1}$ with settings $\mathcal{T}_{m_1}$,
and $\mathcal{C}^{m_1}$ is nothing but the resulting network function.

\paragraph{Early and hybrid fusion cases.}
In these situations, the model can only be written as $\phi(x^{m_1}, x^{m_2})$. 
There is no apparent way to separate the processing of $x^{m_1}$ and $x^{m_2}$ at the architecture level.
In order to mute the influence from $m_2$, we substitute $x^{m_2}$ with a zero vector of the same dimension.
Since the zero vector bears no information about the task, it won't introduce modality competition.
Therefore, one can formally write $\mathcal{E}_{m_1}/m_2 = (\mathbf{0}^{m_2}, \phi, \mathcal{T}, \mathcal{D}_{m_1})$, indicating that the architecture and training settings are the same as for the multi-modal model.
This time $\mathbf{0}^{m_2}$ refers to the zero input of $m_2$ feature components~\footnote{
We also try to use the random inputs for $\mathbf{0}^m$. Our results suggest that there is no big difference between these two implementations. Please refer to the supplementary material for the detailed sanity check of the definition of the mono-modal concept.
}.
Practically, to construct $\mathcal{C}^{m_1}$, we need to \emph{train} $\phi$ on $\mathcal{D}':=\mathcal{D}_{m_1}\otimes\{\mathbf{0}^{m_2}\}$ with $\mathcal{T}$. 
Samples in $\mathcal{D}'$ are of form $(x^{m_1}, \mathbf{0}^{m_2})$.

\subsubsection{Competition strength}

With the mono-modal concepts as a reference, we are ready to quantify the deviation of the multi-modal model responses from those competition-less states.
A linear probing method~\cite{mcgrathAcquisitionChessKnowledge2022} is employed to estimate this deviation.
Specifically, let $z$ be the latent feature before the last classifier layer in the multi-modal model, we train a linear predictor from $z$ to the targeting mono-modal concept $\mathcal{C}^m$,
\begin{equation}
    f^{m}(z) = \mathbf{W} z + \mathbf{b},
\end{equation}
whose parameters $\mathbf{W}$ and $\mathbf{b}$ are determined by minimizing the empirical mean square error of the predictions,
\begin{equation}
\begin{split}
    \mathbf{W}^{m,*}, \mathbf{b}^{m,*}=&\underset{\mathbf{w}, b}{\arg\min} \frac{1}{N}\sum_{i\in[N]} \left\|f^m(z_i) -\mathcal{C}^{m}(x^m_i)\right\|_{2}^{2}\\
    &+\lambda\left(\|\mathbf{W}\|_{2}+\|\mathbf{b}\|_{2}\right),
\end{split}
\end{equation}
where $\left\|\cdot\right\|_{p}$ denotes the $L_p$ norm, $i$ refers to the index of data samples and $\lambda$ is the regularization strength. The $L_2$ regularization term is introduced to avoid overfitting.

The quality of the above linear fitting reflects how much the multi-modal features deviate from their competition-less states. Thus we define the competition strength as
\begin{equation}
    d^{m} = \frac{\sum_{i}\left(\mathcal{C}^{m}(x^m_i) - f^m(z_{i})\right)^2}{ \sum_{i}(\mathcal{C}^{m}\left(x^m_i) - \overline{\mathcal{C}^{m}}\right)^{2}},
\end{equation}
where $\overline{\mathcal{C}^{m}}$ is the mean mono-modal concept value over data samples.
$d^{m}$ measures the quality of the linear predictions with respect to the naive baseline, i.e., simply predicting the mean value. Its value ranges from $0$ to $1$, indicating the weakest and strongest competition levels respectively.

In practice, we reserve two hold-out datasets for computing the competition strength. One of them is used to train the linear predictor and the other to calculate $d^{m}$.

\section{Experiments and discussion}

\subsection{Experimental settings}

In this paper, 
we systematically apply our adaptive gradient modulation approach to situations that
cover different fusion strategies, different modality combinations, and different network architectures.
For the late fusion case, our approach is compared with existing modulation methods.
Moreover, we also include the mono-modal accuracy and the modality competition strength for all the situations.

We carry out experiments~\footnote{
To better demonstrate the universal effectiveness of AGM, we further carry out experiments on the Kinetics-Sound~\cite{kay2017kinetics} using both the late fusion and the FiLM~\cite{perez2018film} fusion strategies. These results are included in the supplementary material due to the space limit.
}
on five popular multi-modal datasets. 
The AV-MNIST~\cite{vielzeuf2018centralnet} is collected for a multi-media classification task that involves disturbed images and audio features. The CREMA-D~\cite{cao2014crema} is an audio-visual dataset for speech emotion recognition which consists of six emotional labels. The UR-Funny~\cite{hasan2019ur} is created for humor detection, involving words (text), gestures (vision), and prosodic cues (acoustic) modalities. The AVE~\cite{tian2018audio} is devised for an audio-visual event localization classification task, including 28 event classes. The CMU-MOSEI~\cite{zadeh2018multimodal} is collected for sentence-level emotion recognition and sentiment analysis, including audio, visual, and text modalities. Here we only use text and audio modalities.

The experiments can be grouped into two classes.  
The first one concerns the performance of our approach and the behavior of modality competition in the late and early fusion strategies across different multi-modal datasets.
We adopt a unified design of the multi-modal models in this class.
The fusion module in the early fusion case is all built with the MAXOUT~\cite{goodfellow2013maxout} network.
In addition, for each dataset, the network models for both fusion strategies use the same encoder architecture.
Specifically, for the AV-MNIST, the CREMA-D, and the Kinetics-Sound datasets, ResNet18~\cite{He_2016_CVPR} is used as an encoder for both the audio and visual modalities. For the UR-Funny dataset, we use Transformer~\cite{vaswani2017attention} for the encoder for all three modalities.

In the second class, we carry out experiments with current SOTA models and show 
that our approach can also enhance more complex models in a realistic application.
For the AVE dataset, the PSP~\cite{zhou2021positive} network is used, which features elaborately designed methods that align the audio and visual representations during fusion.
For the CMU-MOSEI dataset, we adopt the Transformer-based joint-encoding (TBJE) ~\cite{delbrouck2020transformer} as the model. 
TBJE jointly encodes input modalities through the modular co-attention and the glimpse layer.

Our code is implemented in Pytorch 1.2, and experiments are run on a single NVIDIA 3090 GPU. For the detailed experimental settings and hyper-parameters, please refer to the supplementary material.


\begin{table}[!ht]
    \begin{center}
    \setlength{\tabcolsep}{1.20mm}
    \begin{tabular}{c|lccccc}
    \toprule
    \multicolumn{2}{l}{AV-MNIST} & $Acc$ & $Acc_{a}$ & $Acc_{v}$  & $d^{a}$ & $d^{v}$ \\ 
    \midrule
    \multirow{7}{*}{
        \begin{tabular}{@{}c@{}}
            \rotatebox[origin=c]{90}{\makecell{\small Late fusion}}
        \end{tabular}
    } & $\mathcal{C}^a$ & - & 39.61 & -   & - & - \\ 
      & $\mathcal{C}^v$ & - & - & 65.14   & - & - \\
      & Joint-Train  & 69.77 & 16.05 & 55.83  & 0.7838 & 0.1408 \\
      & G-Blending & 70.32 & 14.36 & 56.59  & 0.7963 & 0.1359 \\
      & Greedy & 70.65 & 18.80 & 63.46  & 0.7358 & 0.1340 \\
      & MSES & 70.68 & 27.50 & 63.34  & 0.7538 & 0.1372 \\
      & MSLR & 70.62 & 22.72 & 62.92  & 0.7300 & 0.1437 \\
      & OGM-GE & 71.08 & 24.53 & 55.85  & 0.7445 & 0.1617 \\
      & AGM & \textbf{72.14} & 38.90 & 63.65 & 0.6787 & 0.1197 \\
    \midrule
    \multirow{4}{*}{
        \begin{tabular}{@{}c@{}}
            \rotatebox[origin=c]{90}{\makecell{\small Early fusion}}
        \end{tabular}
    } & $\mathcal{C}^a$ & - & 41.60 & -  & - & -  \\
      & $\mathcal{C}^v$ & - & - & 65.46   & - & - \\ 
      & Joint-Train & 71.15 & 24.28 & 60.14  & 0.7668 & 0.1825 \\
      & AGM & \textbf{72.26} & 47.79 & 68.48  & 0.7146 & 0.1796 \\
    \bottomrule
    \end{tabular}
    \end{center}
    \caption{Accuracy ($Acc$, $Acc_a$, $Acc_v$) and the competition strength ($d^{a}$, $d^{v}$) on the AV-MNIST dataset for multi-modal models using different fusion strategies. In late fusion, comparison with several modality-specific intervention methods: Modality-Specific Early Stop (MSES), Modality-Specific Learning Rate(MSLR), and On-the-fly Gradient Modulation Generalization Enhancement (OGM-GE). 
    The results of Joint-Train are included as baselines. $\mathcal{C}_a$ and $\mathcal{C}_v$ indicate the performance of audio and visual modality concepts, respectively. The best results are shown in {\bf bold}.}
    \label{tab:AVMNIST}
\end{table}

\begin{table*}[!ht]
    \begin{center}
    \setlength{\tabcolsep}{3mm}
    \begin{tabular}{c|lccccccc}
    \toprule
    \multicolumn{2}{l}{UR-Funny} & $Acc$ & $Acc_a$ & $Acc_v$ & $Acc_t$ &  $d^{a}$ & $d^{v}$ & $d^{t}$ \\ 
    \midrule
    \multirow{7}{*}{
        \begin{tabular}{@{}c@{}}
            \rotatebox[origin=c]{90}{\makecell{\small Late fusion}}
        \end{tabular}
    } & $\mathcal{C}^a$ & - & 59.23 & - & -  &  - & - & - \\
      & $\mathcal{C}^v$ & - & - & 53.16 & -   &  - & - & - \\
      & $\mathcal{C}^t$ & - & - & - & 63.46 & - & - & - \\
      & Joint-Train & 64.50 & 50.31 & 51.53 & 49.78 &  0.5558 & 0.1058 & 0.4513 \\
      & MSES  & 64.23 & 50.31 & 49.69 & 57.87 &  0.5605 & 0.1028 & 0.4592 \\
      & MSLR & 64.74 & 50.31 & 48.62 & 49.69 & 0.5257 & 0.0975 & 0.4316 \\
      & AGM & \textbf{65.97} & 54.87 & 49.36 & 62.22 &  0.5234 & 0.0725 & 0.5147 \\
    \midrule
    \multirow{5}{*}{
        \begin{tabular}{@{}c@{}}
            \rotatebox[origin=c]{90}{\makecell{\small Early fusion}}
        \end{tabular}
    } & $\mathcal{C}^a$ & - & 58.25 & - & - &  - & - & - \\
      & $\mathcal{C}^v$ & - & - & 53.29 & - & - & - & - \\
      & $\mathcal{C}^t$ & - & - & - & 61.07 & - & - & - \\
      & Joint-Train & 65.15 & 54.87 & 50.86 & 54.14 &  0.7217 & 0.2672 & 0.2906 \\
      & AGM & \textbf{66.07} & 64.87 & 55.20 & 63.36 & 0.6962 & 0.2697 & 0.3200 \\
    \bottomrule
    \hline
    \end{tabular}
    \end{center}
    \caption{The same as Table \ref{tab:AVMNIST}, but for UR-Funny dataset. The involved modalities are audio, visual, and text.}
    \label{tab:UR-Funny}
\end{table*}


\begin{table}[!ht]
    \begin{center}
    \setlength{\tabcolsep}{1.2mm}
    \begin{tabular}{c|lccccc}
        \toprule
        \multicolumn{2}{l}{CREMA-D} & $Acc$ & $Acc_a$ & $Acc_v$ & $d^{a}$ & $d^{v}$ \\
        \midrule
    \multirow{9}{*}{
        \begin{tabular}{@{}c@{}}
            \rotatebox[origin=c]{90}{\makecell{\small Late fusion}}
        \end{tabular}
    } & $\mathcal{C}^a$ & - & 62.63 & -  & - & - \\
      & $\mathcal{C}^v$ & - & - & 75.93  & - & - \\
      & Joint-Train & 61.14 & 57.10 & 22.72  & 0.4593 & 0.7555 \\ 
      & G-Blending & 62.03 & 19.58 & 16.89  & 0.4706 & 0.8005 \\
      & Greedy & 63.08 & 43.05 & 16.89 & 0.4598 & 0.7661 \\
      &  MSES & 60.99 & 54.86 & 22.57  & 0.4607 & 0.7546 \\
      & MSLR & 64.42 & 54.86 & 26.31 & 0.4614 & 0.7150 \\
      & OGM-GE & 68.16 & 55.16 & 36.32  & 0.5448 & 0.6929 \\
      & AGM & \textbf{78.48} & 48.58 & 57.85 & 0.6624 & 0.5067  \\
    \midrule
    \multirow{4}{*}{
        \begin{tabular}{@{}c@{}}
            \rotatebox[origin=c]{90}{\makecell{\small Early fusion}}
        \end{tabular}
    } & $\mathcal{C}^a$ & - & 61.29 & -  & - & - \\
      & $\mathcal{C}^v$ & - & - & 75.78 & - & - \\
      & Joint-Train & 61.88 & 42.60 & 16.89  & 0.5345 & 0.9905 \\
      & AGM  & \textbf{81.46} & 76.53 & 80.42  & 0.8753 & 0.6496 \\ 
        \bottomrule
    \end{tabular}
    \end{center}
    \caption{The same as Table \ref{tab:AVMNIST}, but for CREMA-D dataset.}
    \label{tab:CREMAD}
\end{table}

\begin{table}[!ht]
    \begin{center}
    \setlength{\tabcolsep}{1.6mm}
    \begin{tabular}{lccccccc}
        \toprule
        {AVE} & $Acc$ & $Acc_{a}$ & $Acc_{v}$ & $d^{a}$ & $d^{v}$ \\   
        \midrule
        $\mathcal{C}^a$ & - & 65.00 & -  & - & - \\
        $\mathcal{C}^v$ & - & - & 64.69  &  - & - \\
        PSP & 76.02 &  52.58 & 50.18 & 0.6223 & 0.6232 \\
        AGM & \textbf{77.11} & 72.34 & 70.68  & 0.6198 & 0.6337 \\
        \midrule\midrule
        {CMU-MOSEI}  & $Acc$ & $Acc_{t}$ & $Acc_{a}$ & $d^{t}$ & $d^{a}$ \\
        \midrule
        $\mathcal{C}^t$ & - & 80.92 & - & - & - \\
        $\mathcal{C}^a$ & - & - & 74.46 & - & - \\
        TBJE & 80.91 & 73.59 & 73.08 & 0.5794 & 0.9450 \\
        AGM & \textbf{81.76} & 79.41 & 73.08 & 0.5774 & 0.9540 \\
        \bottomrule
    \end{tabular}
    \end{center}
    \caption{Accuracy and competition strength on AVE and MOSEI dataset for the general joint-training network with elaborating fusion structures network. Audio and visual are involved in the AVE dataset and audio and text in MOSEI. PSP stands for general joint training network for the AVE dataset and TBJE for the CMU-MOSEI dataset. $\mathcal{C}_a$, $\mathcal{C}_v$ and $\mathcal{C}_a$ indicate the performance of audio, visual, and text modality, respectively. The best results are shown in {\bf bold}.}
    \label{tab:complexfusion_table}
\end{table}

\subsection{The effectiveness of AGM}

In this subsection, we focus on the $Acc$ column in all the tables and demonstrate the universal effectiveness of our AGM method in improving the model performance.

\Cref{tab:AVMNIST,tab:CREMAD,tab:UR-Funny} summarize the results on the AV-MNIST, the CREMA-D, and the UR-Funny dataset, respectively. 
In the late fusion cases, our approach is compared with the Modality-Specific Early Stopping (MSES) and Modality-Specific Learning Rate (MSLR) methods.
For situations with only two modalities, we also include the results of the Gradient Blending (G-Blending), Characterizing and Overcoming the Greedy Nature of Learning (Greedy), and On-the-fly Gradient Modulation Generalization Enhancement (OGM-GE) method.

It is evident that our approach constantly improves the performance w.r.t. the Joint-Train case and achieves the best accuracy in all situations. 
In the late fusion case, while all modulation methods generally boost the performance compared to the Joint-Train baseline, our approach exceeds the second-best one for a gap of at least $1.06\%$. 
It is notable that the improvement in the early fusion case by our approach is comparable with the ones in late fusion cases.
We note the significant increase in accuracy on CREMA-D, where, after modulating, the results of our approach are $17.34\%$ and $19.58\%$ higher than the ones of Joint-Train in late and early fusion, respectively.
There is also a gap of $10.34\%$ between our approach and OGM-GE.
Such supersizing effectiveness may be attributed to the fact that
the most informative modality in CREMA-D, i.e., the visual modality, is considerably under-exploited in the Joint-Train.
In fact, the mono-modal accuracy of the visual modality is only $22.72\%$, which is much lower than its potential performance of the mono-modal concept, i.e., $75.93\%$.
We observe that the improvement from MSES and MSLR is often very limited. 
Actually, on CREMA-D the accuracy of MSES in the late fusion case is worse than the one of Joint-Train. 
This could be the consequence that MSES only controls the time to stop training and, thus, can only provide limited guidance to the weights update.

We next show that our approach can also boost the performance of existing SOTA models. Those models normally equip with elaborately designed fusion modules to ensure higher prediction accuracy.
\Cref{tab:complexfusion_table} shows the results on the AVE dataset and CMU-MOSEI dataset, on which the improvements are $1.09\%$ and $0.85\%$, respectively.
It is worth noting that all other modulation methods can not apply to such complex situations, as there are no separable branches in the network models for different modalities.

AGM adjusts the modulation coefficients based on the running average of the mono-modal cross entropy which serves as a reference of idea relative strengths of individual modalities. Additional experiments demonstrate that this reference is better than the brutal force requirement of equal contribution from all modalities.
Further, we consider an in-depth comparison between AGM and the OGM-GE as their performance outstands in our experiments.
Specifically, we investigate whether the Generalization Enhancement (GE) technique can hence AGM and, in turn, whether a running average reference can boost the performance of OGM-GE.
We find that neither provides an improvement.
The details of the above-mentioned results can be found in the supplementary material.

Combining all the above results, we conclude that our modulation approach can help boost the model performance regardless of the fusion strategy, the number and types of involved modalities, and the network architecture.

\begin{figure*}
\begin{center}
\includegraphics[width=1.0\linewidth]{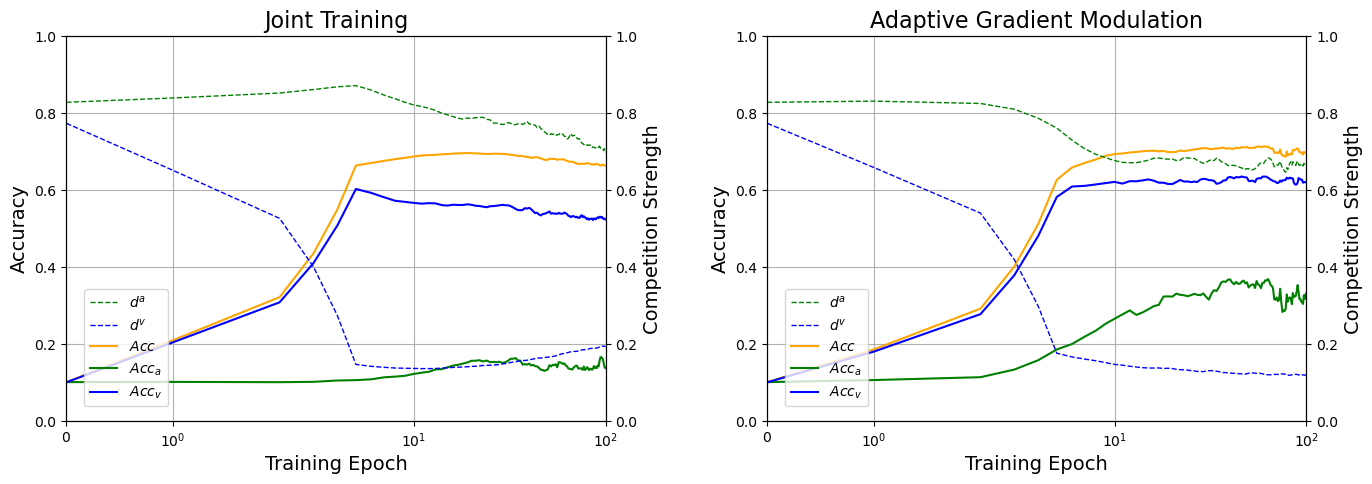}
\end{center}
   \caption{Accuracy ($Acc$, $Acc_a$, $Acc_v$) and competition strength ($d^{a}$, $d^{v}$) of joint-training multimodal model and multimodal model with AGM using addition fusion method on the validation set of the AV-MNIST dataset. The left is the joint-training multimodal model and the right is the multimodal model with our proposed AGM.}
\label{fig:avmnist_normal_vs_gradmod}
\end{figure*}

\subsection{Modality competition}

The competition strength metric provides us a base to analyze the states of individual modalities in a joint-trained model
and understand the mechanism of how the modulation methods work.

In the following, we first compare the changes in competition strength before and after modulating and investigate what is brought to the multi-modal model by our adaptive gradient modulation. This follows a discussion of the modality competition behavior.

\subsubsection{Gradient modulation \& modality competition}

Our primary concern is how the modulation affects the model performance in terms of changing the competition state.
The modality competition directly measures the deviation from the competition-less state and provides more accurate information about the competition state compared to the mono-modal accuracy, which mainly reflects the information in a single modality. 
Generally, we distinguish two different types of change in competition strength.

In the first type, modality competition is mitigated by modulation.
The results on AV-MNIST (~\Cref{tab:AVMNIST}) exemplify this situation.
For both fusion strategies, the competition strengths of audio ($d^a$) and visual ($d^v$) modalities decrease, and their mono-modal accuracy ($Acc_a$ and $Acc_v$) increases as well as the multi-modal performance. 
This suggests that suppressing the competition, allows the model to better utilize inputs from different modalities.
\Cref{fig:avmnist_normal_vs_gradmod} illustrates the change in performance and competition strength along with training.
For the joint training baseline (left panel in~\Cref{fig:avmnist_normal_vs_gradmod}), 
$d^a$ increases while $d^v$ decreases in the initial training stage up to the $9$-th epoch.
Hence, the model initially learns information from the visual modality.
Indeed, $Acc_a$ is almost the random guess while $Acc_v$ is close to the full multi-modal accuracy.
In later epochs, $d^a$ starts to decrease and its mono-modal accuracy increases accordingly.
On the other hand, the increase of $d^v$ is accompanied by the decrease of $Acc_v$.
When adaptive gradient modulation is applied (right panel in~\Cref{fig:avmnist_normal_vs_gradmod}), 
the competition strength of both modalities decreases along training and converges to lower values than their counterpart in the joint training case.
At the same time, their mono-modal accuracy keeps increasing. 
We find that the model starts to learn the audio modality at a relatively earlier epoch and $Acc_a$ is boosted considerably.

In the second type, the competition of some modalities could be strengthened. 
Results in~\Cref{tab:CREMAD,tab:UR-Funny,tab:complexfusion_table} belong to this type.
For CREMA-D, $d^v$ decreases while $d^a$ increases. 
This allows the model to better exploit the visual modality~\footnote{
We remark that, in this case, the modality collapse in joint training on CREMA-D can be attributed to the modality competition. 
}, which is more informative~\footnote{
The accuracy of the visual mono-modal concept is higher than the one of the audio modality.
}. 
Similar behaviors are observed on the AVE and CMU-MOSEI datasets.
In both cases, the modulation leads to a decrease in competition strength of the more informative modality, i.e., the audio modality of AVE and the text modality of CMU-MOSEI.
The results for UR-Funny differ from previous cases.
It mainly reflects a balance in information usage between the audio and text modalities. 
Interestingly, we note that even though the text modality possesses better information, its $d^t$ increases after modulation. 
We suspect this could be due to a high-order effect when multiple modalities are present. In other words, combining the text and the visual modalities could be more informative than combining the audio and visual modalities.

In summary, the results quantitatively demonstrate the behavior behind the effectiveness of our modulation method. In most cases, 
the picture is clear that while the raw model possesses a certain bias towards some modalities, the modulation pushes the model to rely on the more informative modalities~\footnote{
Note that better use of informative modalities does not necessarily lead to low competition strengths of these modalities.
}.

\subsubsection{Behavior of modality competition}

In the following, we proceed to investigate the modality competition in the joint training situation.
We systematically study the competition's behavior from various perspectives that cover the model's preference towards individual modalities, the relation to the fusion strategy, and the relation to the input data.

\paragraph{Existence of preferred modality.}
Our results reveal that modality competition is commonly present in multi-modal models. 
In fact, there is at least one modality with non-trivial competition strength in all situations.
However, we emphasize that it is not necessary for a multi-modal model to have a dominant modality.
The results on AVE (~\Cref{tab:complexfusion_table}) provide such an example.
The balance of the two modalities, in this case, could be attributed to the elaborately designed fusion method in the PSP model. 
In addition, we recognize a trend in all the experiments that the modality with the lowest competition strength always has a higher mono-modal accuracy. This suggests that there exists the model-preferred modality, which the raw multi-modal model tends to explore.
This preference will be broken by the modulation which encourages more efficient usage of modality information.

\paragraph{Relation to fusion strategy.}
The modality competition strengths are similar in the late and early fusion cases. For example, 
in~\Cref{tab:UR-Funny} for the UR-Funny dataset, audio modality is always with the strongest competition, the text modality the second, and the visual modality the weakest.
Other results show similar behavior. 
As this tendency is independent of the fusion strategy, our results suggest that the strength of competition may depend more on the task and the input data.

\paragraph{Relation to modality information.}
It is intuitive to expect that the modality with higher information for the task will have lower competition strength, i.e., being better exploited by the model.
However, it is not always the case.
While the above intuition applies to the results on AV-MNIST and CREMA-D datasets, the visual modality in CREMA-D is under-explored in the joint training case even though it is more informative.
Moreover, for the UR-Funny dataset, the visual modality, which contains less information, has a very low competition strength in the joint training case. 
In conclusion, current results do not support any correlation between the modality information and the competition strength.

\paragraph{Relation to modality type.}
To study whether the modality type affects the competition states, we compare the results of CREMA-D and AV-MNIST.
Both datasets are composed of visual and audio modalities, and the visual modality is more informative.
In addition, our experiments on these two datasets share the same network architecture.
Nonetheless, the competition state of the visual modality in CREMA-D is opposite to the one in AV-MNIST.
Therefore, the strength of modality competition tends to be unrelated to the modality type.

\section*{Acknowledgments}

This work was supported by the National Science and Technology Innovation 2030 Project of China (Nos. 2021ZD0202600) and the National Natural Science Foundation of China (NSFC) (Nos. U22B2063).

\nocite{langley00}

\bibliography{agm}

\begin{thebibliography}{34}
\providecommand{\natexlab}[1]{#1}
\providecommand{\url}[1]{\texttt{#1}}
\expandafter\ifx\csname urlstyle\endcsname\relax
  \providecommand{\doi}[1]{doi: #1}\else
  \providecommand{\doi}{doi: \begingroup \urlstyle{rm}\Url}\fi

\bibitem[Allen-Zhu \& Li(2020)Allen-Zhu and Li]{allen2020towards}
Allen-Zhu, Z. and Li, Y.
\newblock Towards understanding ensemble, knowledge distillation and
  self-distillation in deep learning.
\newblock \emph{arXiv preprint arXiv:2012.09816}, 2020.

\bibitem[Antol et~al.(2015)Antol, Agrawal, Lu, Mitchell, Batra, Zitnick, and
  Parikh]{antol2015vqa}
Antol, S., Agrawal, A., Lu, J., Mitchell, M., Batra, D., Zitnick, C.~L., and
  Parikh, D.
\newblock Vqa: Visual question answering.
\newblock In \emph{Proceedings of the IEEE international conference on computer
  vision}, pp.\  2425--2433, 2015.

\bibitem[Baltrušaitis et~al.()Baltrušaitis, Ahuja, and
  Morency]{baltrusaitisMultimodalMachineLearning2017}
Baltrušaitis, T., Ahuja, C., and Morency, L.-P.
\newblock Multimodal machine learning: A survey and taxonomy.
\newblock URL \url{http://arxiv.org/abs/1705.09406}.

\bibitem[Cao et~al.(2014)Cao, Cooper, Keutmann, Gur, Nenkova, and
  Verma]{cao2014crema}
Cao, H., Cooper, D.~G., Keutmann, M.~K., Gur, R.~C., Nenkova, A., and Verma, R.
\newblock Crema-d: Crowd-sourced emotional multimodal actors dataset.
\newblock \emph{IEEE transactions on affective computing}, 5\penalty0
  (4):\penalty0 377--390, 2014.

\bibitem[Chen et~al.(2020)Chen, Xie, Vedaldi, and Zisserman]{chen2020vggsound}
Chen, H., Xie, W., Vedaldi, A., and Zisserman, A.
\newblock Vggsound: A large-scale audio-visual dataset.
\newblock In \emph{ICASSP 2020-2020 IEEE International Conference on Acoustics,
  Speech and Signal Processing (ICASSP)}, pp.\  721--725. IEEE, 2020.

\bibitem[Delbrouck et~al.(2020)Delbrouck, Tits, Brousmiche, and
  Dupont]{delbrouck2020transformer}
Delbrouck, J.-B., Tits, N., Brousmiche, M., and Dupont, S.
\newblock A transformer-based joint-encoding for emotion recognition and
  sentiment analysis.
\newblock \emph{arXiv preprint arXiv:2006.15955}, 2020.

\bibitem[Fujimori et~al.(2020)Fujimori, Endo, Kawai, and
  Mochizuki]{fujimoriModalitySpecificLearningRate2020}
Fujimori, N., Endo, R., Kawai, Y., and Mochizuki, T.
\newblock Modality-{{Specific Learning Rate Control}} for {{Multimodal
  Classification}}.
\newblock In Palaiahnakote, S., {Sanniti di Baja}, G., Wang, L., and Yan, W.~Q.
  (eds.), \emph{Pattern {{Recognition}}}, Lecture {{Notes}} in {{Computer
  Science}}, pp.\  412--422, {Cham}, 2020. {Springer International Publishing}.
\newblock ISBN 978-3-030-41299-9.
\newblock \doi{10.1007/978-3-030-41299-9_32}.

\bibitem[Gat et~al.(2021)Gat, Schwartz, and
  Schwing]{gatPerceptualScoreWhat2021a}
Gat, I., Schwartz, I., and Schwing, A.
\newblock Perceptual {{Score}}: {{What Data Modalities Does Your Model
  Perceive}}?, October 2021.

\bibitem[Geng et~al.(2021)Geng, Han, Zhang, and Hu]{geng2021uncertainty}
Geng, Y., Han, Z., Zhang, C., and Hu, Q.
\newblock Uncertainty-aware multi-view representation learning.
\newblock In \emph{Proceedings of the AAAI Conference on Artificial
  Intelligence}, volume~35, pp.\  7545--7553, 2021.

\bibitem[Goodfellow et~al.(2013)Goodfellow, Warde-Farley, Mirza, Courville, and
  Bengio]{goodfellow2013maxout}
Goodfellow, I., Warde-Farley, D., Mirza, M., Courville, A., and Bengio, Y.
\newblock Maxout networks.
\newblock In \emph{International conference on machine learning}, pp.\
  1319--1327. PMLR, 2013.

\bibitem[Han et~al.(2022)Han, Zhang, Fu, and Zhou]{han2022trusted}
Han, Z., Zhang, C., Fu, H., and Zhou, J.~T.
\newblock Trusted multi-view classification with dynamic evidential fusion.
\newblock \emph{IEEE transactions on pattern analysis and machine
  intelligence}, 45\penalty0 (2):\penalty0 2551--2566, 2022.

\bibitem[Hasan et~al.(2019)Hasan, Rahman, Zadeh, Zhong, Tanveer, Morency,
  et~al.]{hasan2019ur}
Hasan, M.~K., Rahman, W., Zadeh, A., Zhong, J., Tanveer, M.~I., Morency, L.-P.,
  et~al.
\newblock Ur-funny: A multimodal language dataset for understanding humor.
\newblock \emph{arXiv preprint arXiv:1904.06618}, 2019.

\bibitem[He et~al.(2016)He, Zhang, Ren, and Sun]{He_2016_CVPR}
He, K., Zhang, X., Ren, S., and Sun, J.
\newblock Deep residual learning for image recognition.
\newblock In \emph{Proceedings of the IEEE Conference on Computer Vision and
  Pattern Recognition (CVPR)}, June 2016.

\bibitem[Hessel \& Lee(2020)Hessel and Lee]{hesselDoesMyMultimodal2020a}
Hessel, J. and Lee, L.
\newblock Does my multimodal model learn cross-modal interactions? {{It}}'s
  harder to tell than you might think!, October 2020.

\bibitem[Hu et~al.(2022)Hu, Li, and Zhou]{huSHAPEUnifiedApproach2022}
Hu, P., Li, X., and Zhou, Y.
\newblock {{SHAPE}}: {{An Unified Approach}} to {{Evaluate}} the
  {{Contribution}} and {{Cooperation}} of {{Individual Modalities}}.
\newblock In \emph{Proceedings of the {{Thirty-First International Joint
  Conference}} on {{Artificial Intelligence}}}, pp.\  3064--3070, {Vienna,
  Austria}, July 2022. {International Joint Conferences on Artificial
  Intelligence Organization}.
\newblock ISBN 978-1-956792-00-3.
\newblock \doi{10.24963/ijcai.2022/425}.

\bibitem[Huang et~al.(2022)Huang, Lin, Zhou, Yang, and
  Huang]{huangModalityCompetitionWhat2022}
Huang, Y., Lin, J., Zhou, C., Yang, H., and Huang, L.
\newblock Modality {{Competition}}: {{What Makes Joint Training}} of
  {{Multi-modal Network Fail}} in {{Deep Learning}}? ({{Provably}}).
\newblock In \emph{Proceedings of the 39th {{International Conference}} on
  {{Machine Learning}}}, pp.\  9226--9259. {PMLR}, June 2022.

\bibitem[Ilievski \& Feng(2017)Ilievski and Feng]{ilievski2017multimodal}
Ilievski, I. and Feng, J.
\newblock Multimodal learning and reasoning for visual question answering.
\newblock \emph{Advances in neural information processing systems}, 30, 2017.

\bibitem[Kay et~al.(2017)Kay, Carreira, Simonyan, Zhang, Hillier,
  Vijayanarasimhan, Viola, Green, Back, Natsev, et~al.]{kay2017kinetics}
Kay, W., Carreira, J., Simonyan, K., Zhang, B., Hillier, C., Vijayanarasimhan,
  S., Viola, F., Green, T., Back, T., Natsev, P., et~al.
\newblock The kinetics human action video dataset.
\newblock \emph{arXiv preprint arXiv:1705.06950}, 2017.

\bibitem[Liang et~al.(2021)Liang, Lyu, Fan, Wu, Cheng, Wu, Chen, Wu, Lee, Zhu,
  et~al.]{liang2021multibench}
Liang, P.~P., Lyu, Y., Fan, X., Wu, Z., Cheng, Y., Wu, J., Chen, L., Wu, P.,
  Lee, M.~A., Zhu, Y., et~al.
\newblock Multibench: Multiscale benchmarks for multimodal representation
  learning.
\newblock \emph{arXiv preprint arXiv:2107.07502}, 2021.

\bibitem[Ma et~al.(2022)Ma, Ren, Zhao, Testuggine, and Peng]{ma2022multimodal}
Ma, M., Ren, J., Zhao, L., Testuggine, D., and Peng, X.
\newblock Are multimodal transformers robust to missing modality?
\newblock In \emph{Proceedings of the IEEE/CVF Conference on Computer Vision
  and Pattern Recognition}, pp.\  18177--18186, 2022.

\bibitem[McGrath et~al.(2022)McGrath, Kapishnikov, Toma{\v s}ev, Pearce,
  Hassabis, Kim, Paquet, and Kramnik]{mcgrathAcquisitionChessKnowledge2022}
McGrath, T., Kapishnikov, A., Toma{\v s}ev, N., Pearce, A., Hassabis, D., Kim,
  B., Paquet, U., and Kramnik, V.
\newblock Acquisition of {{Chess Knowledge}} in {{AlphaZero}}, August 2022.

\bibitem[Peng et~al.(2022)Peng, Wei, Deng, Wang, and
  Hu]{pengBalancedMultimodalLearning2022a}
Peng, X., Wei, Y., Deng, A., Wang, D., and Hu, D.
\newblock Balanced {{Multimodal Learning}} via {{On-the-fly Gradient
  Modulation}}.
\newblock In \emph{2022 {{IEEE}}/{{CVF Conference}} on {{Computer Vision}} and
  {{Pattern Recognition}} ({{CVPR}})}, pp.\  8228--8237, {New Orleans, LA,
  USA}, June 2022. {IEEE}.
\newblock ISBN 978-1-66546-946-3.
\newblock \doi{10.1109/CVPR52688.2022.00806}.

\bibitem[Perez et~al.(2018)Perez, Strub, De~Vries, Dumoulin, and
  Courville]{perez2018film}
Perez, E., Strub, F., De~Vries, H., Dumoulin, V., and Courville, A.
\newblock Film: Visual reasoning with a general conditioning layer.
\newblock In \emph{Proceedings of the AAAI conference on artificial
  intelligence}, volume~32, 2018.

\bibitem[Shrikumar et~al.(2018)Shrikumar, Su, and
  Kundaje]{shrikumarComputationallyEfficientMeasures2018b}
Shrikumar, A., Su, J., and Kundaje, A.
\newblock Computationally {{Efficient Measures}} of {{Internal Neuron
  Importance}}, July 2018.

\bibitem[Tian et~al.(2018)Tian, Shi, Li, Duan, and Xu]{tian2018audio}
Tian, Y., Shi, J., Li, B., Duan, Z., and Xu, C.
\newblock Audio-visual event localization in unconstrained videos.
\newblock In \emph{Proceedings of the European Conference on Computer Vision
  (ECCV)}, pp.\  247--263, 2018.

\bibitem[Vaswani et~al.(2017)Vaswani, Shazeer, Parmar, Uszkoreit, Jones, Gomez,
  Kaiser, and Polosukhin]{vaswani2017attention}
Vaswani, A., Shazeer, N., Parmar, N., Uszkoreit, J., Jones, L., Gomez, A.~N.,
  Kaiser, {\L}., and Polosukhin, I.
\newblock Attention is all you need.
\newblock \emph{Advances in neural information processing systems}, 30, 2017.

\bibitem[Vielzeuf et~al.(2018)Vielzeuf, Lechervy, Pateux, and
  Jurie]{vielzeuf2018centralnet}
Vielzeuf, V., Lechervy, A., Pateux, S., and Jurie, F.
\newblock Centralnet: a multilayer approach for multimodal fusion.
\newblock In \emph{Proceedings of the European Conference on Computer Vision
  (ECCV) Workshops}, pp.\  0--0, 2018.

\bibitem[Wang et~al.(2020{\natexlab{a}})Wang, Tran, and Feiszli]{wang2020makes}
Wang, W., Tran, D., and Feiszli, M.
\newblock What makes training multi-modal classification networks hard?
\newblock In \emph{Proceedings of the IEEE/CVF conference on computer vision
  and pattern recognition}, pp.\  12695--12705, 2020{\natexlab{a}}.

\bibitem[Wang et~al.(2020{\natexlab{b}})Wang, Tran, and
  Feiszli]{wangWhatMakesTraining2020}
Wang, W., Tran, D., and Feiszli, M.
\newblock What {{Makes Training Multi-Modal Classification Networks Hard}}?
\newblock In \emph{2020 {{IEEE}}/{{CVF Conference}} on {{Computer Vision}} and
  {{Pattern Recognition}} ({{CVPR}})}, pp.\  12692--12702, {Seattle, WA, USA},
  June 2020{\natexlab{b}}. {IEEE}.
\newblock ISBN 978-1-72817-168-5.
\newblock \doi{10.1109/CVPR42600.2020.01271}.

\bibitem[Wu et~al.(2021)Wu, Yu, Chen, Tenenbaum, and Gan]{wu2021star}
Wu, B., Yu, S., Chen, Z., Tenenbaum, J.~B., and Gan, C.
\newblock Star: A benchmark for situated reasoning in real-world videos.
\newblock In \emph{Thirty-fifth Conference on Neural Information Processing
  Systems Datasets and Benchmarks Track (Round 2)}, 2021.

\bibitem[Wu et~al.(2022)Wu, Jastrzebski, Cho, and Geras]{wu2022characterizing}
Wu, N., Jastrzebski, S., Cho, K., and Geras, K.~J.
\newblock Characterizing and overcoming the greedy nature of learning in
  multi-modal deep neural networks.
\newblock In \emph{International Conference on Machine Learning}, pp.\
  24043--24055. PMLR, 2022.

\bibitem[Yao \& Mihalcea(2022)Yao and
  Mihalcea]{yaoModalityspecificLearningRates2022}
Yao, Y. and Mihalcea, R.
\newblock Modality-specific {{Learning Rates}} for {{Effective Multimodal
  Additive Late-fusion}}.
\newblock In \emph{Findings of the {{Association}} for {{Computational
  Linguistics}}: {{ACL}} 2022}, pp.\  1824--1834, {Dublin, Ireland}, May 2022.
  {Association for Computational Linguistics}.
\newblock \doi{10.18653/v1/2022.findings-acl.143}.

\bibitem[Zadeh et~al.(2018)Zadeh, Liang, Poria, Cambria, and
  Morency]{zadeh2018multimodal}
Zadeh, A.~B., Liang, P.~P., Poria, S., Cambria, E., and Morency, L.-P.
\newblock Multimodal language analysis in the wild: Cmu-mosei dataset and
  interpretable dynamic fusion graph.
\newblock In \emph{Proceedings of the 56th Annual Meeting of the Association
  for Computational Linguistics (Volume 1: Long Papers)}, pp.\  2236--2246,
  2018.

\bibitem[Zhou et~al.(2021)Zhou, Zheng, Zhong, Hao, and Wang]{zhou2021positive}
Zhou, J., Zheng, L., Zhong, Y., Hao, S., and Wang, M.
\newblock Positive sample propagation along the audio-visual event line.
\newblock In \emph{Proceedings of the IEEE/CVF Conference on Computer Vision
  and Pattern Recognition}, pp.\  8436--8444, 2021.

\end{thebibliography}
\bibliographystyle{icml2023}

\newpage
\appendix
\twocolumn

\section{Experiment Details}
\setcounter{table}{4}
\subsection{Datasets}
\paragraph{AV-MNIST~\cite{vielzeuf2018centralnet}.} The dataset is collected for multi-media classification tasks by assembling visual and audio features. The first modality, disturbed image, is made of the $28\times 28$ PCA-projected MNIST images. The second modality, audio, is made of audio samples on $112\times 122$ spectrograms. The whole dataset includes $70,000$ samples, and the division of the training set and validation set is $6/1$. We randomly selected $10\%$ samples from the training set and validation set to create a development set.

\paragraph{UR-Funny~\cite{hasan2019ur}.} The dataset is created for affective computing tasks that detect humor by the usage of words (text), gestures(vision), and prosodic cues (acoustic). This dataset is collected from the TED talks and uses an equal number of binary labels for each sample. In the experiments, the split of the dataset follows\cite{liang2021multibench}.

\paragraph{CREMA-D~\cite{cao2014crema}.} The dataset is devised for speech emotion recognition with facial and vocal emotional expressions. This dataset contains $6$ most usual emotions: angry, happy, sad, neutral, discarding, disgust, and fear. The whole dataset is randomly divided into $6,027$-sample training set and $669$-sample validation set, as well as $745$-sample testing set.

\paragraph{AVE~\cite{tian2018audio}.} The dataset is an $Audio$-$Visual$ $Event$ (AVE) dataset for audio-visual event localization. This dataset consists of $4,143$ ten-second video clips and has $28$ event classes for each clip together with frame-level annotations. All videos are collected from YouTube. In the experiments, we follow~\cite{tian2018audio} in splitting and pre-processing the dataset.

\paragraph{CMU-MOSEI~\cite{zadeh2018multimodal}.} This dataset is collected for sentence-level sentiment analysis and emotion recognition, containing $23,454$ movie review clips with more than $65.9$ hours of YouTube video by $1,000$ speakers. In our experiments, we only use text and audio modalities, and the train/valid/test set is split into $16,327/1,871/4,662$ samples, respectively.

\paragraph{Kinetics-Sound~\cite{kay2017kinetics}.} The dataset is a multi-modal dataset for human action recognition in videos. The original dataset contains 400 human action classes with at least 400 video clips for each class. In our experiments, we randomly select 30 classes, of which the number of classes is close to OGM-GE~\cite{pengBalancedMultimodalLearning2022a}. This dataset contains 25956 video clips (21545 training, 1494 validation, 2917 test).

\subsection{Implementation details}
For the AV-MNIST dataset, we use ResNet18-based networks as the audio and visual encoders. Following~\cite{chen2020vggsound}, we reduce the number of input channels from $3$ to $1$. 
For the UR-Funny dataset, we use a 4-layer Transformer as the encoder for each modality. The number of attention heads is $8$ and the hidden dimension is $768$.
In the experiments on the above two datasets,
models are trained using the 
SGD optimizer with a $0.9$ momentum and a 1e-4 weight decay.
The initial learning rate is 1e-4, and it decays with a rate of $0.9$ every $70$ epochs.
The batch size is set to $64$.

For the CREMA-D and Kinetics-Sound dataset, we follow the experimental settings used in OGM-GE~\cite{pengBalancedMultimodalLearning2022a}, except for the CREMA-D decay rate in the learning rate scheduler.
This decay rate is now set to $0.9$ to make our training more stable.

For the AVE and CMU-MOSEI datasets, we adopt the same experimental settings in~\cite{zhou2021positive} and~\cite{zadeh2018multimodal}, respectively.

The linear predictor in Section 3.2.2 is implemented with the \verb|sklearn| package.
Specifically, we use ridge regression with the regularization strength $\lambda=120$ for all the situations. The value of $\lambda$ is chosen so that the competition strength converges on the validation sets across all the datasets.

In all the experiments in the main text, the random seed is set to $999$ for reproducibility.


\begin{table}[!ht]
    \begin{center}
    \setlength{\tabcolsep}{1.2mm}
    \begin{tabular}{c|lccccc}
    \toprule
    \multicolumn{2}{l}{AV-MNIST} & $Acc$ & $Acc_{a}$ & $Acc_{v}$  & $d^{a}$ & $d^{v}$ \\
    \midrule
    \multirow{3}{*}{
        \begin{tabular}{@{}c@{}}
            \rotatebox[origin=c]{90}{\makecell{\small zero-pad}}
        \end{tabular}
    }    & $\mathcal{C}^{a}$ & - & 41.60 & - & - & -  \\
         & $\mathcal{C}^{v}$ & -  & - & 65.46  & - & - \\
         & Joint-Train & 71.15 & 24.28 & 60.14 & 0.7668 & 0.1825 \\
    \midrule
    \multirow{3}{*}{
        \begin{tabular}{@{}c@{}}
             \rotatebox[origin=c]{90}{\makecell{\small rand-pad}}
        \end{tabular}
    }   & $\mathcal{C}^{a}$ & - & 40.63 & - & - & - \\
        & $\mathcal{C}^{v}$ & - & - & 65.26 & - & - \\
        & Joint-Train & 71.15 & 24.28 & 60.14 & 0.7147 & 0.2324 \\
    \bottomrule
    \end{tabular}
    \end{center}
    \caption{Comparing the impact of mono-modal concept with different padding methods on competition strength in the AV-MNIST dataset early fusion joint-training. $\textbf{zero-pad}$ indicates padding the input modality with zero vector and $\textbf{rand-pad}$ pad input modality with normal distribution. }
    \label{tab:robust}
\end{table}

\section{Sanity Check} 
In this section, we justify the definition of the proposed competition strength metric.
As linear probing is a standard technique, we are mostly concerned about the robustness of the mono-modal concept.

To this end, 
We first train the mono-modal concept with different random seeds in initialization on the AV-MNIST dataset. The result is shown in~\Cref{tab:stability}. As expected, corresponding competition strengths are of similar magnitudes.

We then compare the cases where the mono-modal concepts are computed using different padding methods.
Recall that we have adopted zero-padding for $\mathbf{0}^m$ to represent the absence of the modality $m$. In this control experiment, we use the random-padding instead. 
In other words, all the elements in $\mathbf{0}^m$ are drawn independently from the normal distribution $N(0,1)$. 
It is arguable that both the zero-padding and random-padding stand for the competition-less state as they carry no task-relevant information.
Note that the padding method only matters in the early and hybrid fusion cases.
\Cref{tab:robust} summarises the results on the AV-MNIST dataset with early fusion models. 
Clearly, the values of competition strength in the zero-padding case are close to the corresponding ones in the random-padding case.

At last, we compare the performance of the mono-modal concept in different fusion strategies.
Recall that the mono-modal concept is a function that maps the mono-modal input to a vector in $\mathbb{R}^K$, which can be used for prediction. 
The performance of the mono-modal concept refers to its prediction accuracy and, hence, 
represents the amount of task-relevant information in the corresponding modality.
From the results in Table $1$ to $3$, we find that 
the performance of the mono-modal concept is very similar in the late and early fusion cases on each dataset.
It is noteworthy that the performance of mono-model concepts in~\Cref{tab:stability,tab:robust} are all close to each other as well. 
This is desirable since the amount of task-relevant information should be independent of specific models.

In summary, the results verify the robustness of the mono-modal concept under different situations and indicate that the competition strength is a well-defined metric.

\section{Additional Results}

\begin{table}[!ht]
    \begin{center}
    \setlength{\tabcolsep}{1.1mm}
    \begin{tabular}{c|lccccc}
        \toprule
        \multicolumn{2}{l}{Kinetics-Sound} & $Acc$ & $Acc_{a}$ & $Acc_{v}$ & $d^{a}$ & $d^{v}$ \\   
        \midrule
        \multirow{4}{*}{
        \begin{tabular}{@{}c@{}}
            \rotatebox[origin=c]{90}{\makecell{\small Late Fusion}}
        \end{tabular}
        }
        & $\mathcal{C}^a$ & - & 42.06 & -  & - & - \\
        & $\mathcal{C}^v$ & - & - & 49.23  & - & - \\
        & Joint-Train & 52.78 & 39.92 & 23.84  & 0.6392 & 0.7064 \\ 
        & AGM & 56.93 & 31.01 & 37.04 & 0.7726 & 0.5916 \\
        \midrule 
        \multirow{4}{*}{
        \begin{tabular}{@{}c@{}}
            \rotatebox[origin=c]{90}{\makecell{\small FilM}}
        \end{tabular}
        }
        & $\mathcal{C}^a$ & - & 41.86 & -  & - & - \\
        & $\mathcal{C}^v$ & - & - & 48.76  & - & - \\
        & Joint-Train & 51.17 & 34.76 & 25.32  & 0.6416 & 0.6691 \\ 
        & AGM & 55.73 & 48.56 & 51.57 & 0.6861 & 0.5045 \\
        \bottomrule
    \end{tabular}
    \end{center}
    \caption{Experiments on the Kinetics-Sound dataset with late fusion and FiLM~\cite{perez2018film} strategies. }
    \label{tab:KS_table}
\end{table}
In this section, 
we present additional experiment results on the Kinetics-Sound dataset with both the later fusion and the FiLM fusion~\cite{perez2018film} strategies. 
Apart from the implementation of the fusion module for the FiLM case, the encoder network and training parameters are the same as those in the AV-MNIST late fusion setting.

\Cref{tab:KS_table}
shows the result on the Kinetics-Sound dataset with late fusion and FiLM, the improvement on which are $3.15\%$ and $3.56\%$, respectively.  Comparing joint-train and AGM, the competition strengths of the visual modality decrease for both fusion strategies, which demonstrates that AGM pushes the model to rely on the more informative modality. 
These additional results further demonstrate the universal effectiveness of AGM.

\section{Ablation Study}

\begin{table}[!ht]
    \begin{center}
    \footnotesize
    \setlength{\tabcolsep}{0.9mm}
    \begin{tabular}{c|c|lccccc}
    
    \toprule
    
    \multicolumn{2}{l}{{}} & {} & $Acc$ & $Acc_{a}$ & $Acc_{v}$  & $d^{a}$ & $d^{v}$ \\ 
    \midrule
    \multirow{5}{*}{
        \begin{tabular}{@{}c@{}}
            \rotatebox[origin=c]{90}{\makecell{\small AV-MNIST}}
        \end{tabular}
    } &
    \multirow{3}{*}{
        \begin{tabular}{@{}c@{}}
            \rotatebox[origin=c]{90}{\makecell{\small L-f}}
        \end{tabular}
    } & OGM-GE(RA) & 70.43 & 18.81 & 55.87 & 0.7329 & 0.1362 \\
    &  & AGM(\textbf{1})& 71.63 & 38.35 & 63.50  & 0.6849 & 0.1313 \\
    &  & AGM-GE & 72.03 & 40.24 & 64.52 & 0.7006 & 0.1215 \\
    \cmidrule{2-8}
    
    &
    \multirow{2}{*}{
        \begin{tabular}{@{}c@{}}
            \rotatebox[origin=c]{90}{\makecell{\small E-f}}
        \end{tabular}
    }  & AGM(\textbf{1}) & 71.72 & 67.89 & 66.53 & 0.7640 & 0.1813 \\
    &  & AGM-GE & 71.88 & 35.88 & 67.89  & 0.7368 & 0.1798 \\
    \midrule
    
    \multirow{5}{*}{
        \begin{tabular}{@{}c@{}}
            \rotatebox[origin=c]{90}{\makecell{\small CREMA-D}}
        \end{tabular}
    } &
    \multirow{3}{*}{
        \begin{tabular}{@{}c@{}}
            \rotatebox[origin=c]{90}{\makecell{\small L-f}}
        \end{tabular}
    } & OGM-GE(RA) & 64.28 & 60.69 & 25.41 & 0.4436 & 0.7423 \\
    &  & AGM(\textbf{1}) & 72.05 & 39.46 & 44.39  & 0.6370 & 0.6103 \\
    &  & AGM-GE & 78.03 & 45.44 & 50.22 & 0.6254 & 0.5152   \\
    \cmidrule(lr){2-8}
    &
    \multirow{2}{*}{
        \begin{tabular}{@{}c@{}}
            \rotatebox[origin=c]{90}{\makecell{\small E-f}}
        \end{tabular}
    } & AGM(\textbf{1}) & 71.15 & 69.66 & 73.24 & 0.6507 & 0.6726 \\
    &  & AGM-GE  & 81.02 & 75.49 & 77.73 & 0.8421 & 0.7583 \\ 
    \bottomrule
    \end{tabular}
    \end{center}
    \caption{Experiments on AV-MNIST and CREMA-D with different ablation experiments. OGM-GE(RA) indicates the OGM-GE method discrepancy ratio toward the running average. AGM(1) is our AGM method tunning toward 1. AGM-GE is our AGM with Generalization Enhancement(GE).}
    \label{tab:ablation}
\end{table}
In this section, we provide an in-depth comparison between AGM and OGM-GE as their performance outstands in our experiments. 
Specifically, we tune the AGM discrepancy ratio towards 1 instead of the running average to justify the usefulness of the running average as the reference. 
On the other hand, we try to tune the discrepancy ratio in OGM-GE toward the running average instead of simply 1 to see whether it could improve the performance.
We also integrate our AGM with the generalization enhancement (GE) technique in OGM-GE and run additional experiments to test its comparability with our modulation method.

~\Cref{tab:ablation} shows the result of the above-mentioned experiments on the AV-MNIST and CREMA-D datasets.
The running average of AGM tuning toward 1 improves the performance compared to the joint-training case while being worse than the one using the running average.
It reflects that the running average push model to use the modality with more information. 
We find that the running average does not improve the OGM-GE method, which attributes to that AGM and OGM-GE adopt different ways to compute the discrepancy ratio, the latter may not be compatible with the running average.
Unlike OGM-GE, GE does not improve our AGM. One possible reason is that the running average introduces additional fluctuations in the gradient which is similar to the effect of the noise term in GE.
GE improves the OGM with large performance, but it does not improve our AGM methods. One possible reason is that the running average introduces additional fluctuations in the gradient which is similar to the effect of the noise term in GE.



\end{document}